%% file: root.tex
\documentclass[letterpaper, 10 pt, conference]{ieeeconf}  % Comment this line out if you need a4paper

\IEEEoverridecommandlockouts                              % This command is only needed if 
\usepackage{shortcuts}

\newlength{\capvspace}
\newlength{\figvspace}
\title{\LARGE \bf
GN-Net: The Gauss-Newton Loss for Multi-Weather Relocalization
}

\author{Lukas von Stumberg\textsuperscript{1,2*} \ \ \
Patrick Wenzel\textsuperscript{1,2*} \ \ \
Qadeer Khan\textsuperscript{1,2} \ \ \
Daniel Cremers\textsuperscript{1,2}% <-this % stops a space
\thanks{\textsuperscript{*}These authors contributed equally.}
\thanks{\textsuperscript{1}Technical University of Munich}
\thanks{\textsuperscript{2}Artisense}
}

\begin{document}
\maketitle
\thispagestyle{empty}
\pagestyle{empty}

%%%%%%%%% ABSTRACT
\input{sections/abstract.tex}

%%%%%%%%% BODY TEXT
\input{sections/introduction.tex}
\input{sections/related_work.tex}
\input{sections/method.tex}
\input{sections/benchmark.tex}
\input{sections/experiments.tex}
\input{sections/conclusion.tex}

\bibliographystyle{IEEEtran}
\bibliography{root}  % .bib 

\end{document}

%% file: sections/abstract.tex
\begin{abstract}
Direct SLAM methods have shown exceptional performance on odometry tasks. However, they are susceptible to dynamic lighting and weather changes while also suffering from a bad initialization on large baselines. To overcome this, we propose GN-Net: a network optimized with the novel Gauss-Newton loss for training weather invariant deep features, tailored for direct image alignment. Our network can be trained with pixel correspondences between images taken from different sequences. Experiments on both simulated and real-world datasets demonstrate that our approach is more robust against bad initialization, variations in day-time, and weather changes thereby outperforming state-of-the-art direct and indirect methods. Furthermore, we release an evaluation benchmark for relocalization tracking against different types of weather. Our benchmark is available at~\url{https://vision.in.tum.de/gn-net}.
\end{abstract}

%% file: sections/introduction.tex
\section{Introduction}\label{sec:intro}

In recent years, very powerful visual SLAM algorithms have been proposed~\cite{Klein2007ParallelTA, Mur-ArtalIEEETR2015}. In particular, direct visual SLAM methods have shown great performance, outperforming indirect methods on most benchmarks~\cite{engel14eccv, Alismail2016PhotometricBA, engel2016dso}. They directly leverage the brightness data of the sensor to estimate localization and 3D maps rather than extracting a heuristically selected sparse subset of feature points. As a result, they exhibit a boost in precision and robustness.
Nevertheless, compared to indirect methods, direct methods suffer from two major drawbacks:

\begin{enumerate}
    \item Direct methods need a good initialization, making them less robust for large baseline tracking or cameras with a low frame rate.
    \item Direct methods cannot handle changing lighting/weather conditions. In such situations, their advantage of being able to pick up very subtle brightness variations becomes a disadvantage to the more lighting invariant features.
\end{enumerate}

\begin{figure}
    \centering
    \includegraphics[width=\linewidth]{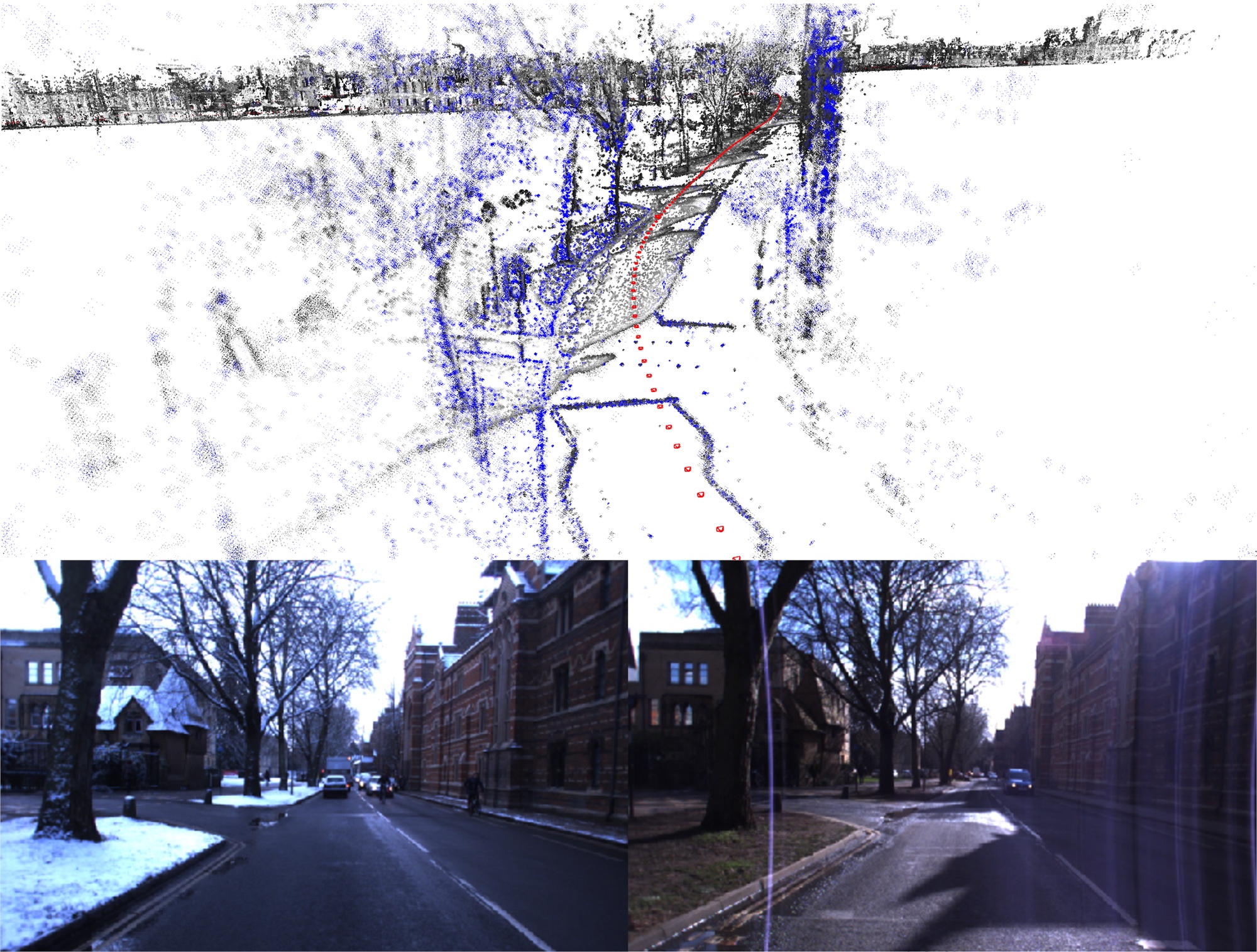}   
    \vspace{\capvspace}
    \caption{We relocalize a snowy sequence from the Oxford RobotCar dataset in a pre-built map created using a sunny weather condition. The points from the prior map (gray) well align with the new points from the current run (blue), indicating that the relocalization is indeed accurate.}
    \label{fig:teaser}
    \vspace{\figvspace}
\end{figure}

In the last years, researchers have tackled the multiple-daytime tracking challenge with deep learning approaches that are designed to convert nighttime images to daytime images \eg using GANs~\cite{LiuNIPS2017, IsolaCVPR2017, PoravICRA2018}. While this improves the robustness to changing lighting, one may ask why images should be the best input representation. Could there be better alternate representations?

This paper addresses the problem of adapting direct SLAM methods to challenging lighting and weather conditions. In this work, we show how to convert images into a multi-dimensional feature map which is invariant to lighting/weather changes and has by construction a larger basin of convergence. Thereby we overcome the aforementioned problems simultaneously. The deep features are trained with a novel Gauss-Newton loss formulation in a self-supervised manner. We employ a Siamese network trained with labels obtained either from simulation data or any state-of-the-art SLAM algorithm. This eliminates the additional cost of human labeling that is typically necessary for training a neural network. We exploit the probabilistic interpretation of the commonly used Gauss-Newton algorithm for direct image alignment. For this, we propose the Gauss-Newton loss which is designed to maximize the probability of identifying the correct pixel correspondence. The proposed loss function thereby enforces a feature representation that is designed to admit a large basin of convergence for the subsequent Gauss-Newton optimization. The superiority of our method stems from its ability to generate these multi-channel, weather-invariant deep features that facilitate relocalization across different weathers. Figure~\ref{fig:teaser} shows how our method can successfully relocalize a snowy sequence in a pre-built map created using a sunny sequence.

In common benchmarks~\cite{SattlerCVPR2018}, localizing accurately in a pre-built map has been tackled by finding nearby images (\eg by using NetVLAD~\cite{ArandjelovicCVPR2016}) and tracking the relative pose (6DOF) between them. However, we propose to split this into two separate tasks. In this work, we focus on the second challenge which we refer to as \emph{relocalization tracking}. This way, we can evaluate its performance in isolation. This is formalized to what we refer to as \emph{relocalization tracking}. Since there is no publicly available dataset to evaluate \emph{relocalization tracking} performance across multiple types of weathers, we are releasing an evaluation benchmark having the following 3 attributes:

\begin{itemize}
    \item It contains sequences from multiple different kinds of weathers.
    \item Pixel-wise correspondences between sequences are provided for both simulated and real-world datasets.
    \item It decouples relocalization tracking from the image retrieval task.
\end{itemize}

The challenge here in comparison with normal pose estimation datasets~\cite{Burri2016Euroc,engel2016monodataset} is that the images involved are usually captured at different daytimes/seasons and there is no good initialization of the pose. 
We summarize the main contributions of our paper as:

\begin{itemize}
    \item We derive the Gauss-Newton loss formulation based on the properties of direct image alignment and demonstrate that it improves the robustness to large baselines and illumination/weather changes. 
    \item Our experimental evaluation shows, that GN-Net outperforms both state-of-the-art direct and indirect SLAM methods on the task of \emph{relocalization tracking}.
    \item We release a new evaluation benchmark for the task of \emph{relocalization tracking} with ground-truth poses. It is collected under dynamic conditions such as illumination changes, and different weathers. Sequences are taken from the the CARLA~\cite{DosovitskiyCoRL2017} simulator as well as from the Oxford RobotCar dataset~\cite{MaddernIJRR2017}. 
\end{itemize}

%% file: sections/related_work.tex
\section{Related Work}\label{sec:related_work}

We review the following main areas of related work: visual SLAM, visual descriptor learning, deep direct image alignment, and image-based relocalization in SLAM. 

\noindent{\textbf{Direct versus indirect SLAM methods: }}Most existing SLAM systems that have used feature descriptors are based on traditional manual feature engineering, such as ORB-SLAM~\cite{Mur-ArtalIEEETR2015}, MonoSLAM~\cite{davison07pami}, and PTAM~\cite{Klein2007ParallelTA}.

An alternative to feature-based methods is to skip the pre-processing step of the raw sensor measurements and rather use the pixel intensities directly.
Popular direct visual methods are DTAM~\cite{newcombe2011iccv}, LSD-SLAM~\cite{engel14eccv}, DSO~\cite{engel2016dso}, and PhotoBundle~\cite{Alismail2016PhotometricBA}. However, the main limitation of direct methods is the \emph{brightness constancy} assumption which is rarely fulfilled in any real-world robotic application~\cite{ParkICRA2017}. The authors of~\cite{AlismailIEEERAL2017} propose to use binary feature descriptors for direct tracking called Bit-planes. While improving the robustness to bad lighting situations it was also found that Bit-planes have a smaller convergence basin than intensities. This makes their method less robust to bad initialization. In contrast, the features we propose \emph{both} improve robustness to lighting and the convergence basin.

\noindent{\textbf{Visual descriptor learning: }} Feature descriptors play an important role in a variety of computer vision tasks. For example,~\cite{ChoyNIPS2016} proposed a novel correspondence contrastive loss which allows for faster training and demonstrates their effectiveness for both geometric and semantic matching across intra-class shape or appearance variations. In~\cite{SchmidtRAL2017}, a deep neural network is trained using a contrastive loss to produce viewpoint- and lighting-invariant descriptors for single-frame localization. The authors of~\cite{RoccoNeurIPS2018} proposed a CNN-based model that learns local patterns for image matching without a global geometric model. \cite{Wohlhart2015LearningDF} uses convolutional neural networks to compute descriptors which allow for efficient detection of poorly textured objects and estimation of their 3D pose. In~\cite{ChangCVPR2017}, the authors propose to train features for optical flow estimation using a Hinge loss based on correspondences. In contrast to our work, their loss function does not have a probabilistic derivation and they do not apply their features to pose estimation.~\cite{GomezOjeda2018LearningBasedIE} uses deep learning to improve SLAM performance in challenging situations. They synthetically create images and choose the one with most gradient information as the ground-truth for training. In contrast to them, we do not limit our network to output images similar to the real world. In~\cite{Czarnowski2017SemanticTF}, the authors compare dense descriptors from a standard CNN, SIFT, and normal image intensities for dense Lucas-Kanade tracking. There, it can be seen that grayscale values have a better convergence basin than the other features, which is something we overcome with our approach.

\noindent{\textbf{Deep direct image alignment: }}BA-NET~\cite{TangarXiv2019} introduces a network architecture to solve the structure from motion (SfM) problem via feature-metric bundle adjustment. Unlike the BA-NET, instead of predicting the depth and the camera motion simultaneously, we propose to only train on correspondences obtained from a direct SLAM system. The advantage is that correspondences are oftentimes easier to obtain than accurate ground-truth poses. Furthermore, we combine our method with a state-of-the-art direct SLAM system and utilize its depth estimation, whereas BA-NET purely relies on deep learning. RegNet~\cite{han2018regnet} is another line of work which tries to replace the handcrafted numerical Jacobian by a learned Jacobian with the help of a depth prediction neural network. However, predicting a dense depth map is often inaccurate and computationally demanding. The authors of~\cite{Lv19cvpr} propose to use a learning-based inverse compositional algorithm for dense image alignment. The drawback of this approach is that the algorithm is very sensitive to the data distribution and constrained towards selecting the right hyperparameters. In~\cite{jaramillo2017direct} they use high-dimensional features in a direct image alignment framework for monocular VO. In contrast to us, they only use already existing features and do not apply them for relocalization.

\noindent{\textbf{Relocalization: }}An important task of relocalization is to approximate the pose of an image by simply querying the most similar image from a database~\cite{fabmap,vlad}. However, this has only limited accuracy unless the 6DOF pose between the queried and the current image is estimated in a second step.
Typically, this works by matching 2D-3D correspondences between an image and a point cloud and estimating the pose using indirect image alignment~\cite{orbslam2}. In contrast, we propose to use direct image alignment paired with deep features.

\noindent{\textbf{Relocalization benchmarks: }} The authors of~\cite{SattlerCVPR2018} have done sequence alignment on the Oxford RobotCar dataset, however, they have not made the matching correspondences public. The Photo Tourism~\cite{SSS:2006} is another dataset providing images and ground-truth correspondences of popular monuments from different camera angles and across different weather/lighting conditions. However, since the images are not recorded as a sequence, relocalization tracking is not possible. Furthermore, their benchmark only supports the submission of features rather than poses, thereby restricting evaluation to only indirect methods. 

%% file: sections/method.tex
\section{Deep Direct SLAM}\label{sec:method}

In this work, we argue that a network trained to output features which produce better inputs for direct SLAM as opposed to normal images should have the following properties:

\begin{itemize}
    \item Pixels corresponding to the same 3D point should have similar features.
    \item Pixels corresponding to different 3D points should have dissimilar features.
    \item When starting in a vicinity around the correct pixel, the Gauss-Newton algorithm should move towards the correct solution.
\end{itemize}

For optimizing the last property, we propose the novel Gauss-Newton loss which makes use of the probabilistic background of the Gauss-Newton algorithm for direct image alignment. The final loss is a weighted sum of the pixel-wise contrastive loss and the Gauss-Newton loss.

\begin{figure}[ht]
    \centering
    \includegraphics[width=\linewidth]{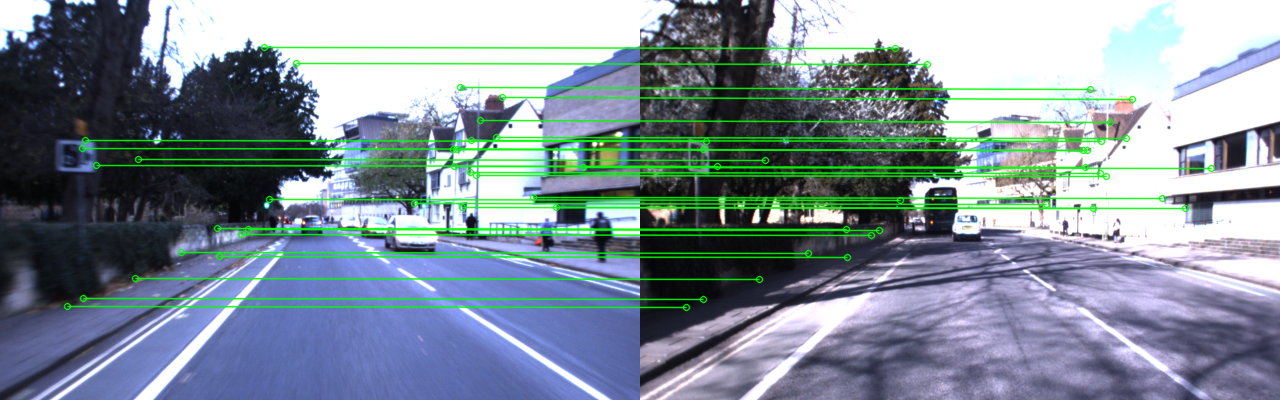}
    \vspace{\capvspace}
    \caption{This figure shows training correspondences between a pair of images from our benchmark.}
    \label{fig:correspondences}
    \vspace{\figvspace}
\end{figure}

\noindent{\bf{Architecture: }}
We are interested in learning a non-linear mapping, which maps images, $\mathbb{R}^{W \times H \times C}$ to a dense visual descriptor space, $\mathbb{R}^{W \times H \times D}$, where each pixel is represented by a $D$-dimensional vector. The training is performed by a Siamese encoder-decoder structured network, where we feed a pair of images, $\mathbf{I}_a$ and $\mathbf{I}_b$, producing multi-scale feature pyramids $\mathbf{F}_a^l$ and $\mathbf{F}_b^l$, where $l$ represents the level of the decoder. For each image pair, we use a certain number of matches, denoted by $N_\text{pos}$, and a certain number of non-matches, denoted by $N_\text{neg}$. A pixel $\mathbf{u}_a \in \mathbb{R}^2$ from image $\mathbf{I}_a$ is considered to be a positive example if the pixel $\mathbf{u}_b \in \mathbb{R}^2$ from image $\mathbf{I}_b$ corresponds to the same 3D vertex (Figure~\ref{fig:correspondences}). We make use of the inherent multi-scale hierarchy of the U-Net~\cite{RonnebergerMICCAI2015} architecture to apply the different loss terms from coarser to finer scaled pyramid levels. With this approach, our learned features will have a larger convergence radius for visual SLAM methods.

\noindent{\bf{Pixelwise contrastive loss: }}
The pixelwise contrastive loss attempts to minimize the distance between positive pairs, and maximize the distance between negative pairs. It can be computed as follows: $\mathcal{L}_{\text{contrastive}}(\mathbf{F}_a, \mathbf{F}_b, l) = \mathcal{L}_{\text{pos}}(\mathbf{F}_a, \mathbf{F}_b, l) + \mathcal{L}_{\text{neg}}(\mathbf{F}_a, \mathbf{F}_b, l)$.

\begin{align}
    \mathcal{L}_{\text{pos}}(\mathbf{F}_a, \mathbf{F}_b, l) = \frac{1}{N_\text{pos}} \, \sum_{N_\text{pos}} D_{\text{feat}}^2 \\
    \mathcal{L}_{\text{neg}}(\mathbf{F}_a, \mathbf{F}_b, l) = \frac{1}{N_\text{neg}} \, \sum_{N_\text{neg}} \max (0, M - D_{\text{feat}})^2
\end{align}

where $D_{\text{feat}}(\cdot)$ is the $L_2$ distance between the feature embeddings: $D_{\text{feat}} = ||\mathbf{F}_a^l(\mathbf{u}_a) - \mathbf{F}_b^l(\mathbf{u}_b)||_2$ and $M$ is the margin and set to 1. 

\noindent{\bf{Gauss-Newton algorithm for direct image alignment: }}
Our learned deep features are ultimately applied to pose estimation. This is done using direct image alignment but generalized to a multi-channel feature map $\mathbf{F}$ with $D$ channels.
The input to this algorithm is a reference feature map $\mathbf{F}$ with known depths for some pixels in the image, and a target feature map $\mathbf{F'}$. The output is the predicted relative pose $\boldsymbol{\xi}$.
Starting from an initial guess the following steps are performed iteratively:

\begin{enumerate}
    \item All points $\mathbf{p}_i$ with known depth values are projected from the reference feature map $\mathbf{F}$ into the target feature map $\mathbf{F'}$ yielding the point $\mathbf{p'}_i$. For each of them a residual vector $\mathbf{r} \in \mathbb{R}^D$ is computed, enforcing that the reference pixel and the target pixel should be similar:
    \begin{align}
        \mathbf{r}_i(\mathbf{p}_i, \mathbf{p'}_i) = \mathbf{F'}(\mathbf{p'}_i) - \mathbf{F}(\mathbf{p}_i)
    \end{align}
    \item For each residual the derivative with respect to the relative pose is:
    \begin{align} \label{eq:jac}
        \mathbf{J}_i = \frac{d\mathbf{r}_i}{d\boldsymbol{\xi}} = \frac{d \mathbf{F'}(\mathbf{p'}_i)}{d\mathbf{p'}_i} \cdot \frac{d\mathbf{p'}_i}{d\boldsymbol{\xi}}
    \end{align}
    Notice that the reference point $\mathbf{p}_i$ does not change for different solutions $\boldsymbol{\xi}$, therefore it does not appear in the derivative.
    \item Using the stacked residual vector $\mathbf{r}$, the stacked Jacobian $\mathbf{J}$, and a diagonal weight matrix $\mathbf{W}$, the Gaussian system and the step $\boldsymbol{\delta}$ is computed as follows:
    \begin{align} \label{eq:step}
        \mathbf{H}=\mathbf{J}^T \mathbf{W} \mathbf{J} \text{~and~} \mathbf{b}= -\mathbf{J}^T\mathbf{W}\mathbf{r} \text{~and~}
        \boldsymbol{\delta} = \mathbf{H}^{-1} \mathbf{b}
    \end{align}
\end{enumerate}

Note that this derivation is equivalent to normal direct image alignment (as done in the frame-to-frame tracking from DSO) when replacing $\mathbf{F}$ with the image $\mathbf{I}$.
In the computation of the Jacobian the numerical derivative of the features $\frac{d\mathbf{F'}(\mathbf{p'}_i)}{d\mathbf{p'}_i}$ is used. As typical images are extremely non-convex this derivative is only valid in a small vicinity (usually 1-2 pixels) around the current solution which is the main reason why direct image alignment needs a good initialization. To partially overcome this, a pyramid scheme is often used.
Usually tracking on multiple channels instead of one can decrease the convergence radius (\cite{AlismailIEEERAL2017, Czarnowski2017SemanticTF}). However, in our case, we train the feature maps to in fact have a larger convergence basin than images by enforcing smoothness in the vicinity of the correct correspondence.

\noindent{\bf{Gauss-Newton on individual pixels: }}
Instead of running the Gauss-Newton algorithm on the 6DOF pose we can instead use it on each point $\mathbf{p}_i$ individually (which is similar to the Lucas-Kanade algorithm~\cite{Lucas1981AnII}). Compared to direct image alignment, this optimization problem has the same residual, but the parameter being optimized is the point position instead of the relative pose.
In this case, the Hessian will be a 2-by-2 matrix and the step $\boldsymbol{\delta}$ can simply be added to the current pixel position (we leave out $\mathbf{W}$ for simplicity): 

\begin{align} \label{eq:2djac}
    \mathbf{J'}_i = \frac{d \mathbf{F'}(\mathbf{p'}_i)}{d\mathbf{p'}_i} \text{ and } \mathbf{H}_i' = \mathbf{J'}_i^T \mathbf{J}_i \text{ and } \mathbf{b'}_i = \mathbf{J'}_i^T \mathbf{r}_i
\end{align}

These individual Gauss-Newton systems can be combined with the system for 6DOF pose estimation (Equation (\ref{eq:step})) using:\newline $
    \mathbf{H} = \sum_i \left(\frac{d\mathbf{p'}_i}{d\boldsymbol{\xi}}\right)^T \mathbf{H'}_i \left( \frac{d\mathbf{p'}_i}{d\boldsymbol{\xi}}\right) \text{ and } \mathbf{b} = \sum_i \left(\frac{d\mathbf{p'}_i}{d\boldsymbol{\xi}}\right)^T \mathbf{b'}_i$
The difference between our simplified systems and the one for pose estimation is only the derivative with respect to the pose, which is much smoother than the image derivative \cite{engel2016dso}.

This means that if the Gauss-Newton algorithm performs well on individual pixels it will also work well on estimating the full pose. Therefore, we propose to train a neural network on correspondences which are easy to obtain, \eg using a SLAM method, and then later apply it for pose estimation. 

We argue that training on these individual points is superior to training on the 6DOF pose. The estimated pose can be correct even if some points contribute very wrong estimates. This increases robustness at runtime but when training we want to improve the information each point provides. Also, when training on the 6DOF pose we only have one supervision signal for each image pair, whereas when training on correspondences we have over a thousand signals. Hence, our method exhibits exceptional generalization capabilities as shown in the results section.

\noindent{\bf{The probabilistic Gauss-Newton loss: }} The linear system described in Equation (\ref{eq:2djac}) defines a 2-dimensional Gaussian probability distribution. The reason is that the Gauss-Newton algorithm tries to find the solution with maximum probability in a least squares fashion. This can be derived using the negative log-likelihood of the Gaussian distribution:
\begin{align}
    E(\mathbf{x}) = -\log f_X(\mathbf{x}) = \\ \frac{1}{2}(\mathbf{x} - \mathbf{\boldsymbol{\mu}})^T \boldsymbol{\Sigma}^{-1} (\mathbf{x} - \boldsymbol{\mu}) + \log\left( 2\pi \sqrt{|\boldsymbol{\Sigma}|} \right) = \\ \frac{1}{2}(\mathbf{x} - \mathbf{\boldsymbol{\mu}})^T \mathbf{H} (\mathbf{x} - \boldsymbol{\mu}) + \log(2\pi) - \frac{1}{2} \log(|\mathbf{H}|) \label{eq:gnloss}
\end{align}
where $\mathbf{x}$ is a pixel position and $\boldsymbol{\mu}$ is the mean.

In the Gauss-Newton algorithm the mean (which also corresponds to the point with maximum probability) is computed with $\boldsymbol{\mu} = \mathbf{x}_s + \boldsymbol{\delta}$, where the $\boldsymbol{\delta}$ comes from Equation~(\ref{eq:step}) and $\mathbf{x}_s$ denotes the start point. To derive this, only the first term is used (because the latter parts are constant for all solutions $\mathbf{x}$). In our case, however, the second term is very relevant, because the network can influence both $\boldsymbol{\mu}$ and $\mathbf{H}$.

This derivation shows, that $\mathbf{H}, \mathbf{b}$ as computed in the GN-algorithm, also define a Gaussian probability distribution with mean $\mathbf{x}_s + \mathbf{H}^{-1} \mathbf{b}$ and covariance $\mathbf{H}^{-1}$.

When starting with an initial solution $\mathbf{x}_s$ the network should assign maximal probability to the pixel that marks the correct correspondence. With $\mathbf{x}$ being the correct correspondence, we therefore use $E(\mathbf{x})$ = Equation~(\ref{eq:gnloss}) as our loss function which we call the \emph{Gauss-Newton loss} (see Algorithm~\ref{alg:gnloss}).

\begin{algorithm}
\caption{Compute Gauss-Newton loss}\label{alg:gnloss}
    \begin{algorithmic}
    \State $\mathbf{F}_a \gets \text{network}(\mathbf{I}_a)$
    \State $\mathbf{F}_b \gets \text{network}(\mathbf{I}_b)$
    \State $e \gets 0$ \Comment{Total error}
    \ForAll{correspondences $\mathbf{u}_a, \mathbf{u}_b$}
        \State $\mathbf{f}_t \gets \mathbf{F}_a(\mathbf{u}_a)$ \Comment{Target feature}
        \State $\mathbf{x}_s \gets \mathbf{u}_b + \text{rand}(\text{vicinity})$ \Comment{Compute start point}
        \State $\mathbf{f}_s \gets \mathbf{F}_b(\mathbf{x}_s)$
        \State $\mathbf{r} \gets \mathbf{f}_s - \mathbf{f}_t$ \Comment{Residual}
        \State $\mathbf{J} \gets \frac{d\mathbf{F}_b}{d\mathbf{x}_s}$ \Comment{Numerical derivative}
        \State $\mathbf{H} \gets \mathbf{J}^T \mathbf{J} + \epsilon \cdot \text{Id}$ \Comment{Added epsilon for invertibility}
        \State $\mathbf{b} \gets \mathbf{J}^T r$
        \State $\boldsymbol{\mu} \gets \mathbf{x}_s - \mathbf{H}^{-1}\mathbf{b}$
        \State $e_1 \gets \frac{1}{2} (\mathbf{u}_b - \boldsymbol{\mu})^T \mathbf{H} (\mathbf{u}_b - \boldsymbol{\mu})$ \Comment{First error term}
        \State $e_2 \gets \log(2 \pi) - \frac{1}{2}  \log(|\mathbf{H}|)$ \Comment{Second error term}
        \State $e \gets e + e_1 + e_2$
    \EndFor
    \end{algorithmic}
\end{algorithm}

In the algorithm, a small number $\epsilon$ is added to the diagonal of the Hessian, to ensure it is invertible. 

\noindent{\bf{Analysis of the Gauss-Newton loss: }}
By minimizing Equation~(\ref{eq:gnloss}) the network has to maximize the probability density of the correct solution. As the integral over the probability densities always has to be 1, the network has the choice to either focus all the density on a small set of solutions (with more risk of being penalized if this solution is wrong), or to distribute the density to more solutions which in turn will have a lower individual density. By maximizing the probability of the correct solution, the network is incentivized to improve the estimated solution and its certainty.

This is also reflected in the two parts of the loss. The first term $e_1 = \frac{1}{2}(\mathbf{u}_b - \mathbf{\boldsymbol{\mu}})^T \mathbf{H} (\mathbf{u}_b - \boldsymbol{\mu})$ penalizes deviations between the estimated and the correct solution, scaled with the Hessian $\mathbf{H}$. The second term $e_2 = \log(2\pi) - \frac{1}{2} \log(|\mathbf{H}|)$ is large if the network does not output enough certainty for its solution. This means that the network can reduce the first error term $e_1$ by making $\mathbf{H}$ smaller. As a consequence, the second error term will be increased, as this will also reduce the determinant of $\mathbf{H}$.
Notice also that this can be done in both dimensions independently. The network has the ability to output a large uncertainty in one direction, but a small uncertainty in the other direction. This is one of the traditional advantages of direct methods which are naturally able to utilize also lines instead of just feature points.

From Equation~(\ref{eq:gnloss}) it can be observed that the predicted uncertainty depends only on the numerical derivative of the target image at the start position. The higher the gradients the higher the predicted certainty. In DSO this is an unwanted effect that is counteracted by the gradient-dependent weighting applied to the cost-function~\cite[Equation (7)]{engel2016dso}. In our case, however, it gives the network the possibility to express its certainty and incentivizes it to output discriminative features.

\begin{figure}
    \centering
    \includegraphics[width=\linewidth]{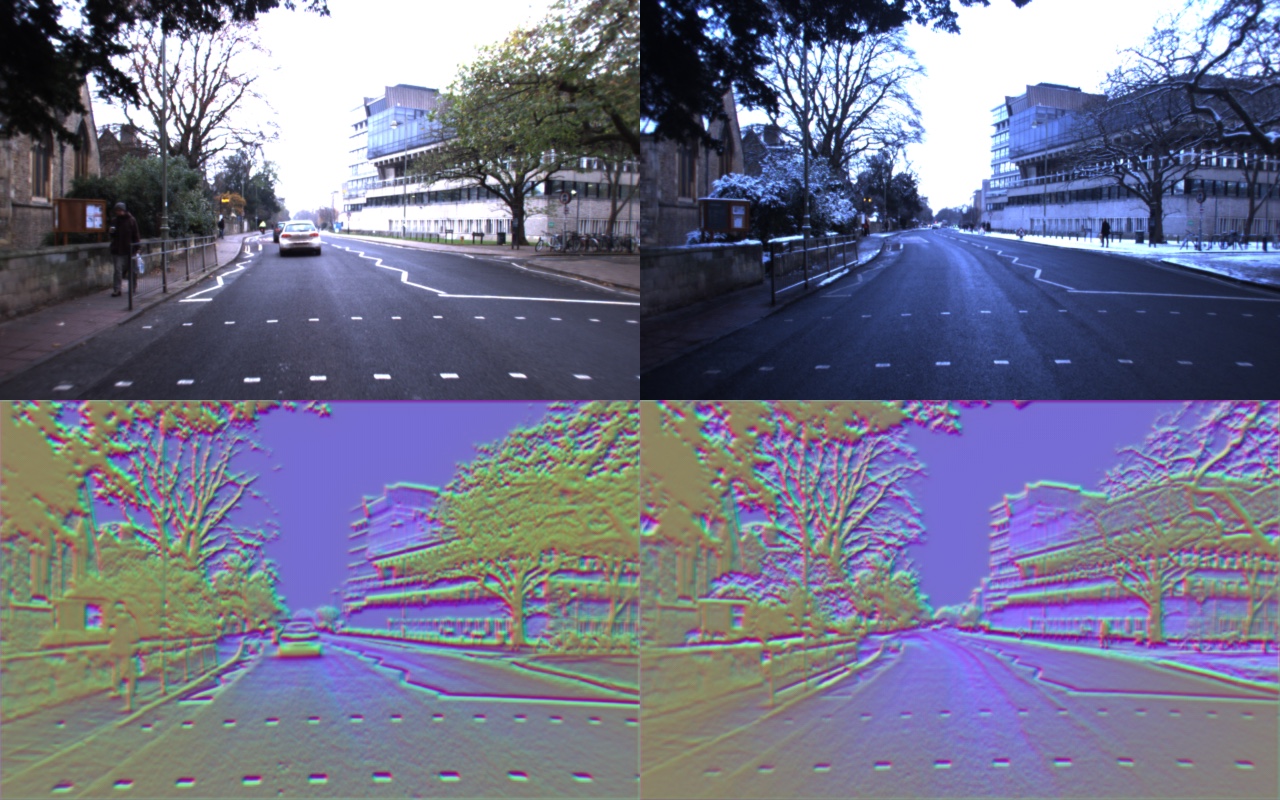}
    \vspace{\capvspace}
    \caption{This figure shows images and their corresponding feature maps predicted by our GN-Net for the Oxford RobotCar dataset. Each column depicts the image and feature map for a sample taken from 2 different sequences. Despite lighting and weather changes, the feature maps are robust to these variations. The visualization of the features shows the high-dimensional descriptors reduced to 3D through PCA.}
    \label{fig:matricesweathers}
    \vspace{\figvspace}
\end{figure}

Upon training the network with our loss formulation, we observe that the features are very similar despite being generated from images taken from sequences with different lighting/weather conditions, as shown in Figure~\ref{fig:matricesweathers}.  

%% file: sections/benchmark.tex
\section{Relocalization Tracking Benchmark}\label{sec:reloctracking}

Previous tasks for localization/odometry can primarily be divided into two categories:

\begin{itemize}
    \item Odometry datasets~\cite{Burri2016Euroc,engel2016monodataset}, where there is a continuous stream of images (sometimes combined with additional sensor data like IMUs).
    \item Image collections where individual images are usually further apart from each other in space/time~\cite{posenet, SattlerCVPR2018}.
\end{itemize}

We argue that for several applications a combination of these two tasks which we refer to as \emph{relocalization tracking} is a more realistic scenario. The idea is that the algorithm has two inputs:

\begin{enumerate}
    \item An image sequence (like a normal odometry dataset).
    \item A collection of individual images (possibly with different weathers/times), each of which shall be tracked against one specific image from point 1.
\end{enumerate}

The algorithm is supposed to track the normal sequential image sequence and at the same time perform tracking of the images in point 2. The advantage of this task is that the used algorithm can utilize the temporally continuous sequence from point 1 to compute accurate depth values for a part of the image (using a standard visual odometry method), which can then be used to improve the tracking of the individual images of point 2. 

This task is very realistic as it comes up when tracking an image sequence and at the same time trying to relocalize this sequence in a prior map. A similar challenge occurs by trying to merge multiple maps from different times. In both cases, one has more information than just a random collection of images. It is important to reiterate here that the task of finding relocalization candidates is not considered but rather tracking them with maximum accuracy/robustness is the focus. This is because our benchmark decouples image retrieval from tracking.

We have created a benchmark for \emph{relocalization tracking} using the CARLA simulator and the Oxford RobotCar dataset. Our benchmark includes ground-truth poses between different sequences for both training, validation, and testing.

\noindent{\bf{CARLA: }}For synthetic evaluations, we use CARLA version 0.8.2. We collect data for 3 different weather conditions representing \emph{WetNoon}, \emph{SoftRainNoon}, and \emph{WetCloudySunset}. We recorded the images at a fixed framerate of 10 frames per second (FPS). At each time step, we record images and its corresponding dense depth map from 6 different cameras with different poses rendered from the simulation engine, which means that the poses in the benchmark are not limited to just 2DOF. The images and the dense depth maps are of size $512 \times 512$. For each weather condition, we collected 3 different sequences comprising 500-time steps with an average distance of 1.6m. This is done for training, validation, and testing, meaning there are 27 sequences, containing 6 cameras each. Training, validation, and test sequences were all recorded in different parts of the CARLA town. We have generated the test sequences after all hyperparameter tuning of our method was finished, meaning we had no access to the test data when developing the method. In accordance, we shall withhold the ground-truth for the test sequences.

\noindent{\bf{Oxford RobotCar: }}Creating a multi-weather benchmark for this dataset imposes various challenges because the GPS-based ground-truth is very inaccurate. To find the relative poses between images from different sequences we have used the following approach. For pairs of images from two different sequences, we accumulate the point cloud captured by the 2D lidar for 60 meters using the visual odometry result provided by the Oxford dataset. The resulting two point clouds are aligned with the global registration followed by ICP alignment using the implementation of Open3D~\cite{open3d}. We provide the first pair of images manually and the following pairs are found using the previous solution. We have performed this alignment for the following sequences: \emph{2014-12-02-15-30-08 (overcast)} and \emph{2015-03-24-13-47-33 (sunny)} for training. For testing, we use the reference sequence \emph{2015-02-24-12-32-19 (sunny)} and align it with the sequences \emph{2015-03-17-11-08-44 (overcast)}, \emph{2014-12-05-11-09-10 (rainy)}, and \emph{2015-02-03-08-45-10 (snow)}. The average relocalization distance across all sequences is 0.84m.

%% file: sections/experiments.tex
\section{Experimental Evaluation}\label{sec:experiments}

We perform our experiments on the \emph{relocalization tracking} benchmark described in Section~\ref{sec:reloctracking}. We demonstrate the multi-weather relocalization performance on both the CARLA and the Oxford RobotCar dataset. For the latter, we show that our method even generalizes well to unseen weather conditions like rain or snow while being trained only on the sunny and overcast conditions. Furthermore, a qualitative relocalization demo\footnote{\label{note1}\url{https://vision.in.tum.de/gn-net}.} on the Oxford RobotCar dataset is provided, where we demonstrate that our GN-Net can facilitate precise relocalization between weather conditions.

We train our method using sparse depths created by running Stereo DSO on the training sequences. We use intra-sequence correspondences calculated using the DSO depths and the DSO pose. Meanwhile, inter-sequence correspondences are obtained using DSO depths and the ground-truth poses provided by our benchmark. The ground truth poses are obtained via Lidar alignment for Oxford and directly from the simulation engine for CARLA as explained in Section~\ref{sec:reloctracking}.
Training is done from scratch with randomly initialized weights and an ADAM optimizer with a learning rate of $10^{-6}$. The image pair fed to the Siamese network is randomly selected from any of the training sequences while ensuring that the images in the pair do not differ by more than 5 keyframes. Each branch of the Siamese network is a modified U-Net architecture with shared weights. Further details of the architecture and training can be found in the supplementary material\textsuperscript{\ref{note1}}. Note that at inference time, only one image is needed to extract the deep visual descriptors, used as input to the SLAM algorithm. While in principle, our approach can be deployed in conjunction with any direct method, we have coupled our deep features with Direct Sparse Odometry (DSO). 

We compare to state-of-the-art \textbf{direct} methods:

\noindent{\bf{Stereo Direct Sparse Odometry (DSO)~\cite{wang2017stereoDSO}:}} Whenever there is a relocalization candidate for a frame we ensure that the system creates the corresponding keyframe. This candidate is tracked using the \emph{coarse tracker}, performing direct image alignment in a pyramid scheme. We use the identity as initialization without any other random guesses for the pose.

\noindent{\bf{GN-Net (Ours):}} Same as with DSO, however, for relocalization tracking, we replace the grayscale images with features created by our GN-Net on all levels of the feature pyramid. The network is trained with the Gauss-Newton loss formulation described in Section~\ref{sec:method}.

We also compare to state-of-the-art \textbf{indirect} methods:

\noindent{\bf{ORB-SLAM~\cite{orbslam2}:}}  
For relocalization tracking, we use the standard feature-based 2-frame pose optimization also used for frame-to-keyframe tracking. We have also tried the RANSAC scheme implemented in ORB-SLAM for relocalization, however, it yielded worse results overall. Thus we will report only the default results.

\noindent{\bf{D2-Net~\cite{DusmanuCVPR2019}, SuperPoint~\cite{DeToneCVPRW2018}:}} For both methods we use the models provided by the authors. The relative pose is estimated using the OpenCV implementation of the PnP algorithm in a RANSAC scheme.

We also evaluated the Deeper Inverse Compositional Algorithm~\cite{Lv19cvpr} on the \emph{relocalization tracking} benchmark. However, the original implementation didn't converge despite multiple training trials with different hyperparameters.

For all our quantitative experiments we plot a cumulative distribution of the relocalization error, which is the norm of the translation between the estimated and the correct solution in meters. For each relocalization error between 0 and 1 meter, it plots the percentage of relocalization candidates that have been tracked with at least this accuracy.

\subsection{Quantitative multi-weather evaluation}
We demonstrate the relocalization tracking accuracy on our new benchmark across different weathers. For these experiments, tracking is performed only across sequences with a different weather condition.

\noindent{\bf{CARLA:}} For this experiment, we train on the training sequences provided by our benchmark. For all learning-based approaches, the best epoch is selected using the relocalization tracking performance on the validation set. The results on the test data are shown in the supplementary\textsuperscript{\ref{note1}}.

\begin{figure*}
  \centering
    \begin{subfigure}{0.32\linewidth}
    \includegraphics[width=\linewidth]{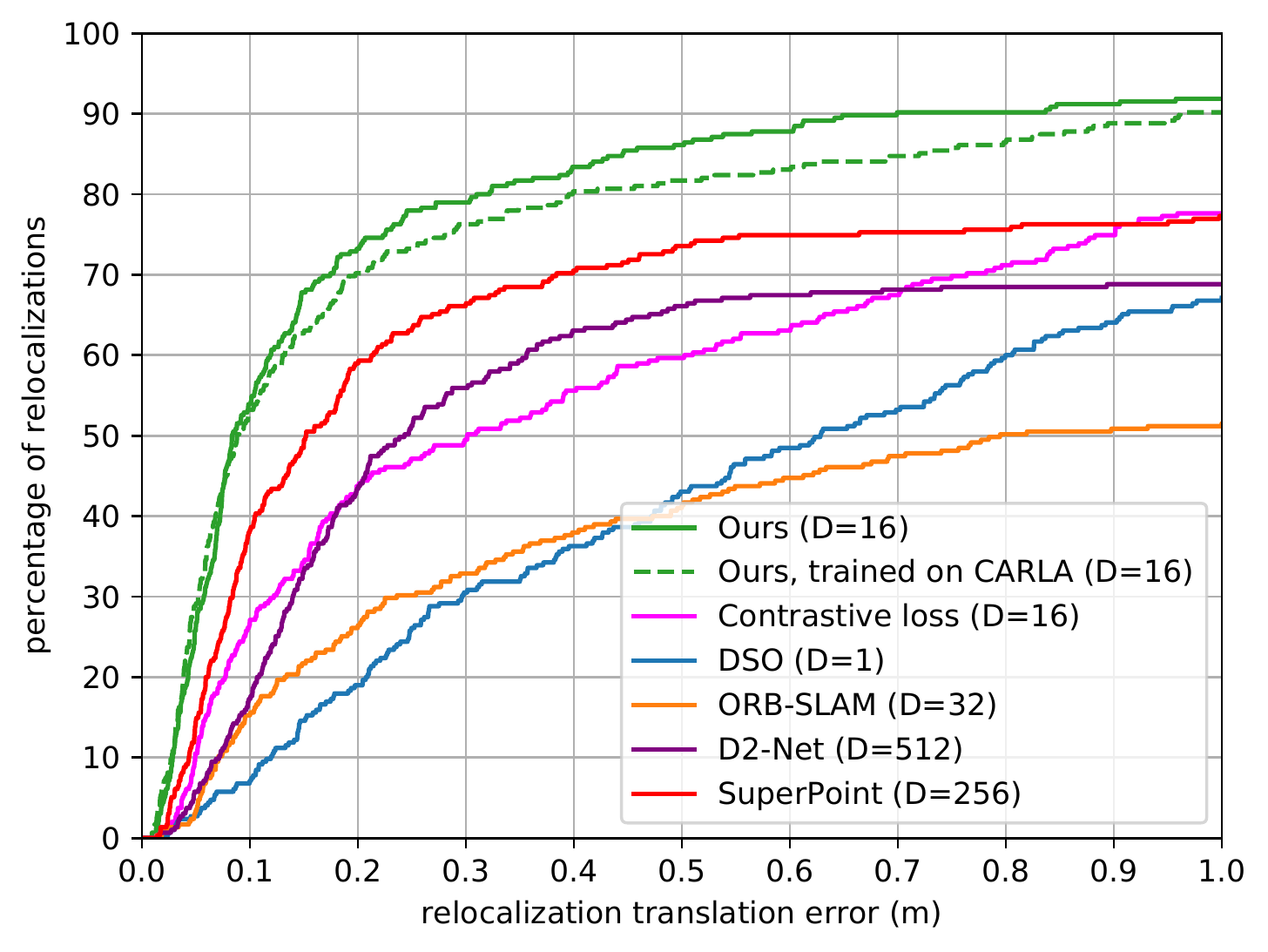}
    \caption{Relocalization sunny and overcast.}
    \label{fig:sunovercast}
  \end{subfigure}
  \begin{subfigure}{0.32\linewidth}
    \includegraphics[width=\linewidth]{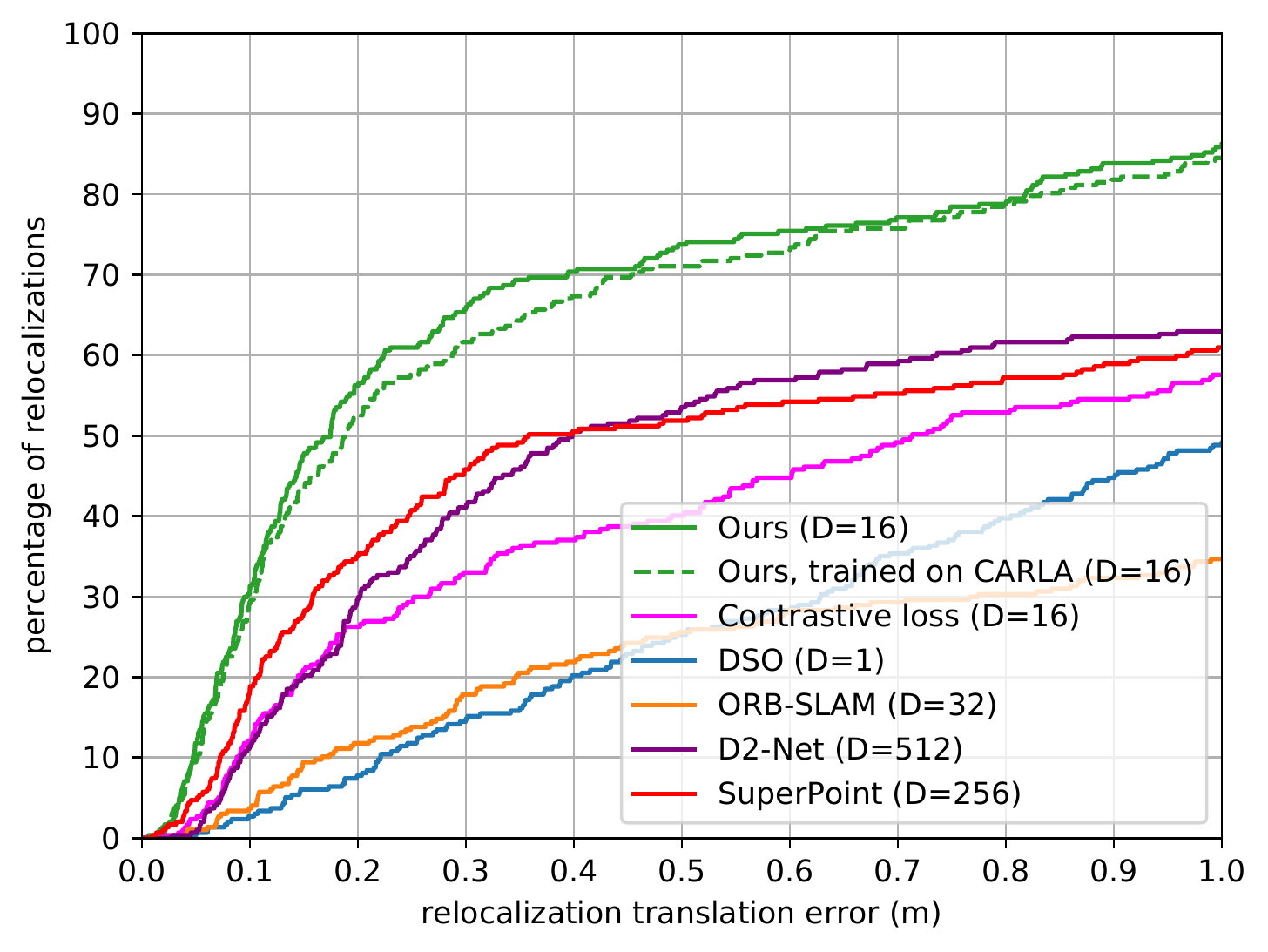}
    \caption{Relocalization sunny and rainy.}
    \label{fig:sunrain}
  \end{subfigure}
   \begin{subfigure}{0.32\linewidth}
    \includegraphics[width=\linewidth]{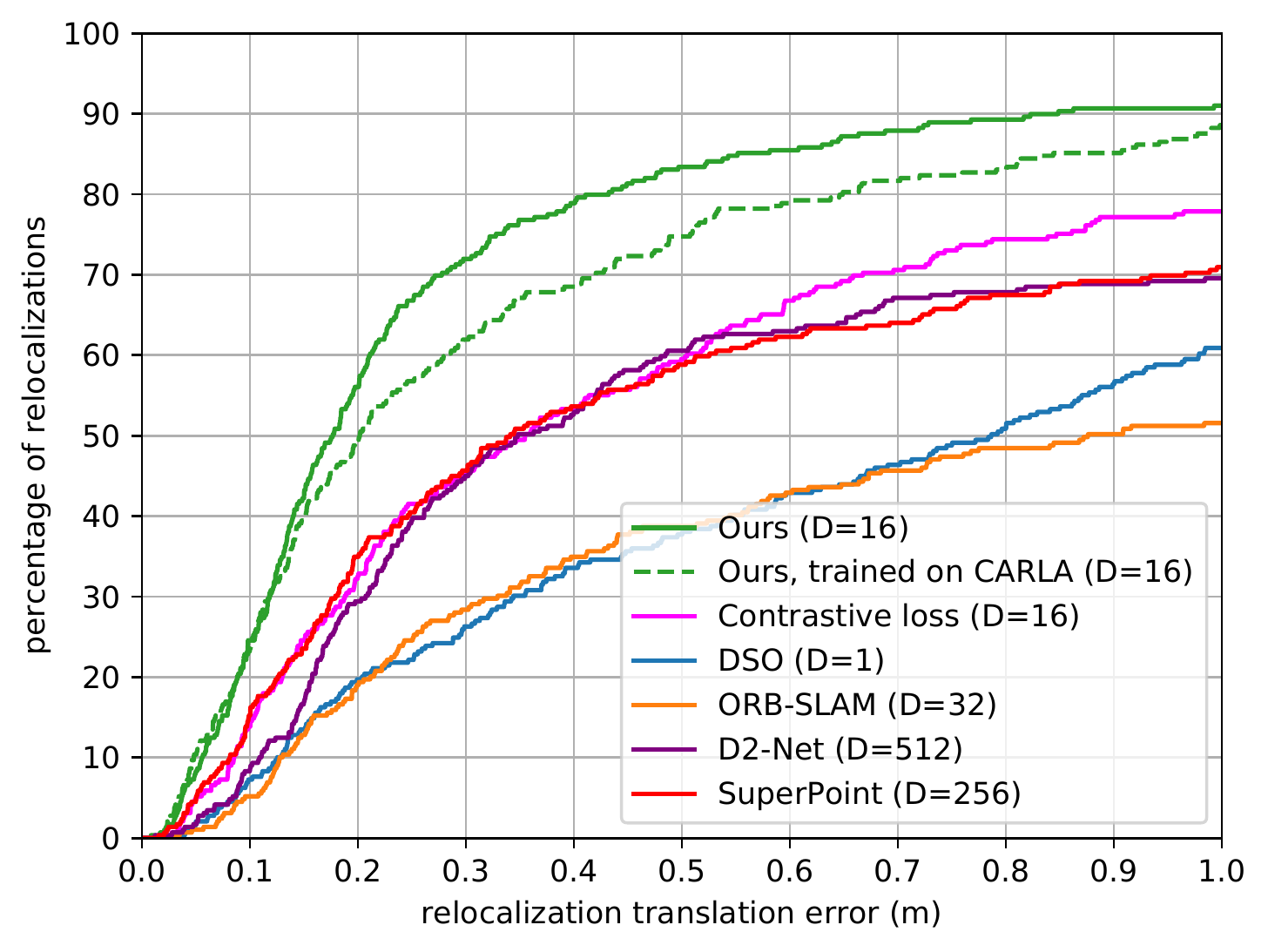}
    \caption{Relocalization sunny and snowy.}
    \label{fig:sunsnow}
  \end{subfigure}
\begin{subfigure}{0.32\linewidth}
    \includegraphics[width=\linewidth]{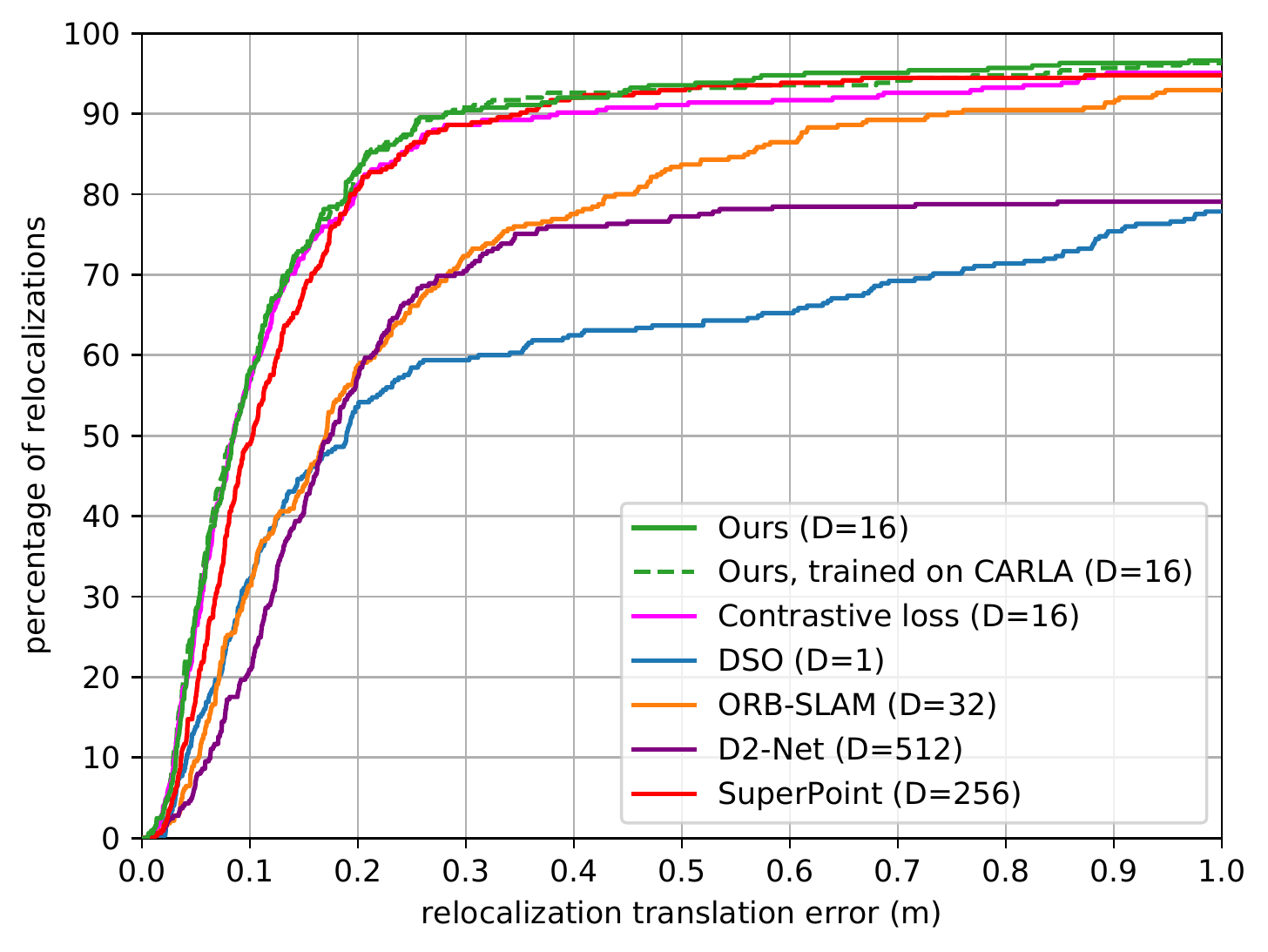}
    \caption{Relocalization overcast and rainy.}
    \label{fig:overcastrain}
  \end{subfigure}
  \begin{subfigure}{0.32\linewidth}
    \includegraphics[width=\linewidth]{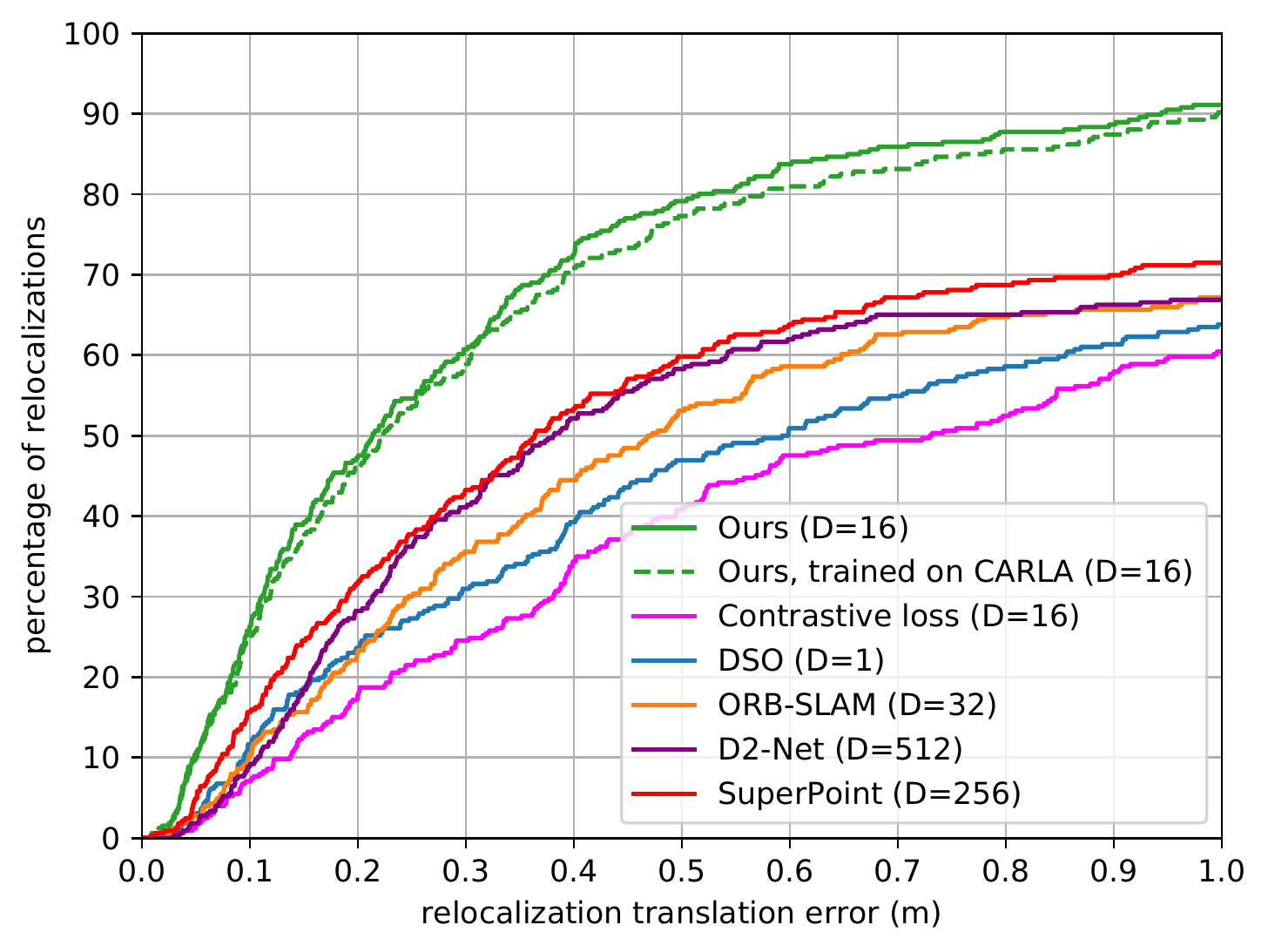}
    \caption{Relocalization overcast and snowy.}
    \label{fig:overcastsnow}
  \end{subfigure}
   \begin{subfigure}{0.32\linewidth}
    \includegraphics[width=\linewidth]{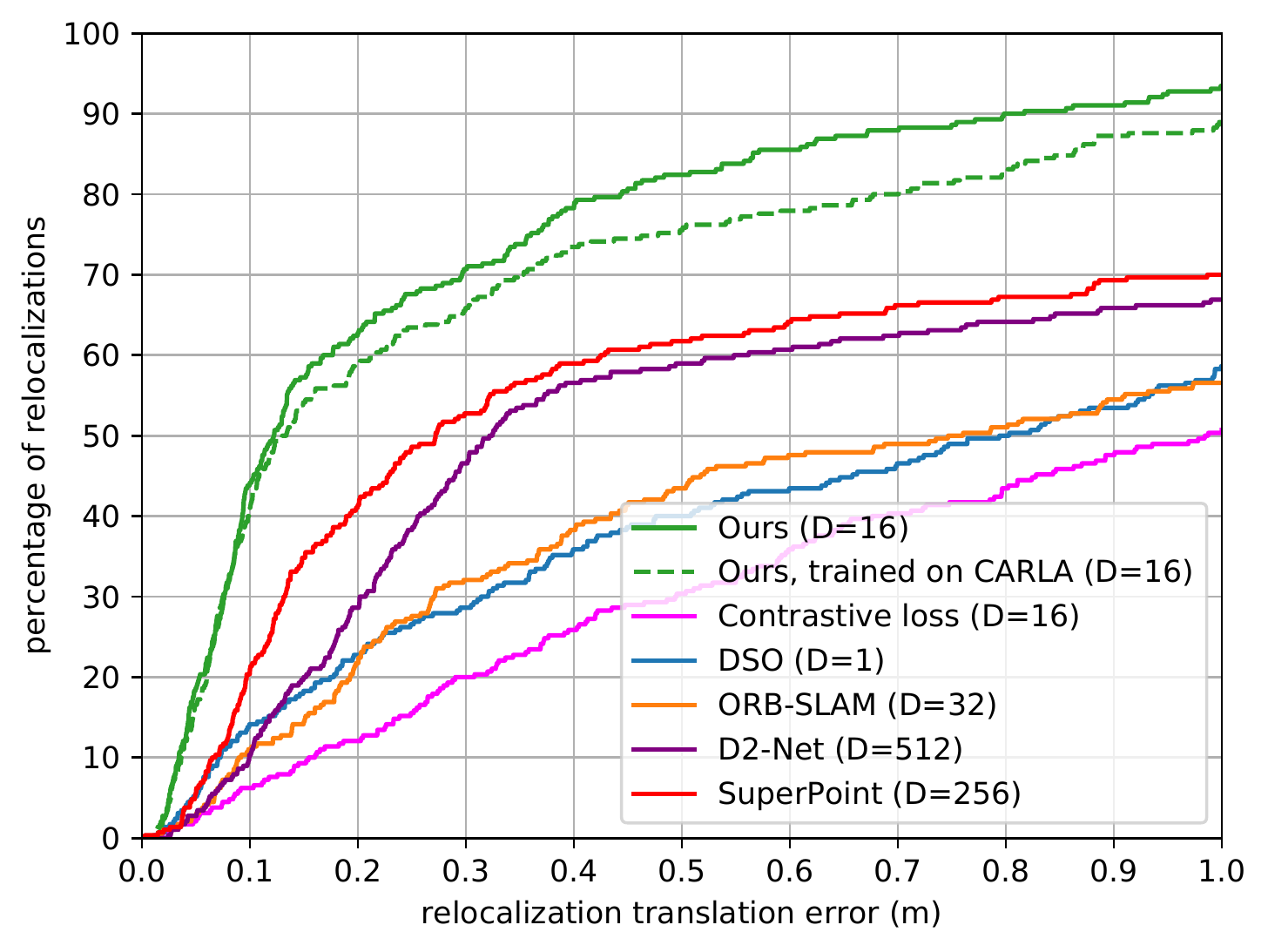}
    \caption{Relocalization rainy and snowy.}
    \label{fig:rainsnow}
  \end{subfigure}
  \caption{This figure shows the cumulative relocalization accuracy on the Oxford RobotCar dataset for different sequences. D denotes the dimension of the feature descriptor. Our method achieves the highest accuracy across all sequences. It is interesting to observe that despite being trained only on two sequences in overcast and sunny condition, our model still generalizes very well to even \emph{unseen} rainy and snowy conditions. Even the model trained only on the synthetic CARLA benchmark outperforms all baselines, showing exceptional generalization capabilities.}
  \label{fig:oxfordgeneralization}
\end{figure*}

\noindent{\bf{Oxford RobotCar:}} We train on the sunny and overcast condition correspondences provided by our \emph{relocalization tracking} benchmark for the Oxford dataset. For the learning-based methods, we select the best epoch based on the relocalization tracking performance on the training set. We use the same hyperparameters that were found using the CARLA validation set. We show the results on the test data in Figure~\ref{fig:oxfordgeneralization}. Our method significantly outperforms the baselines. The Gauss-Newton loss has a large impact as compared to the model trained with only the contrastive loss. 

Figures~\ref{fig:oxfordgeneralization}b-f show how well our model generalizes to unseen weather conditions. Despite being trained only on two sequences with overcast and sunny conditions the results for tracking against a rainy and a snowy sequence are almost the same. Interestingly our model which was trained only on the CARLA benchmark outperforms all baselines significantly.

\subsection{Qualitative multi-weather evaluation}
\begin{figure}[t]
    \centering
    \includegraphics[width=0.95\linewidth]{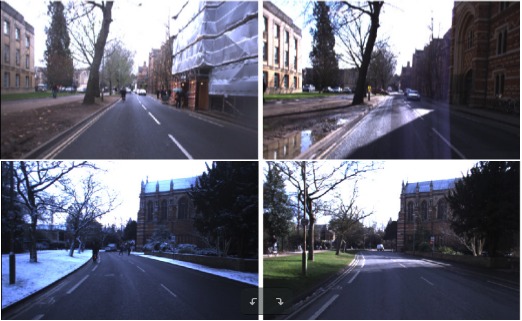}
    %\vspace{\capvspace}
    \caption{shows image pairs used in the qualitative relocalizations. Left: rainy (top row) and snowy (bottom row) images relocalized against the sunny reference images (right).}
    \label{fig:sampleimages}
    \vspace{\figvspace}
\end{figure}

\begin{figure*}[ht]
    \centering
    \includegraphics[width=0.95\linewidth]{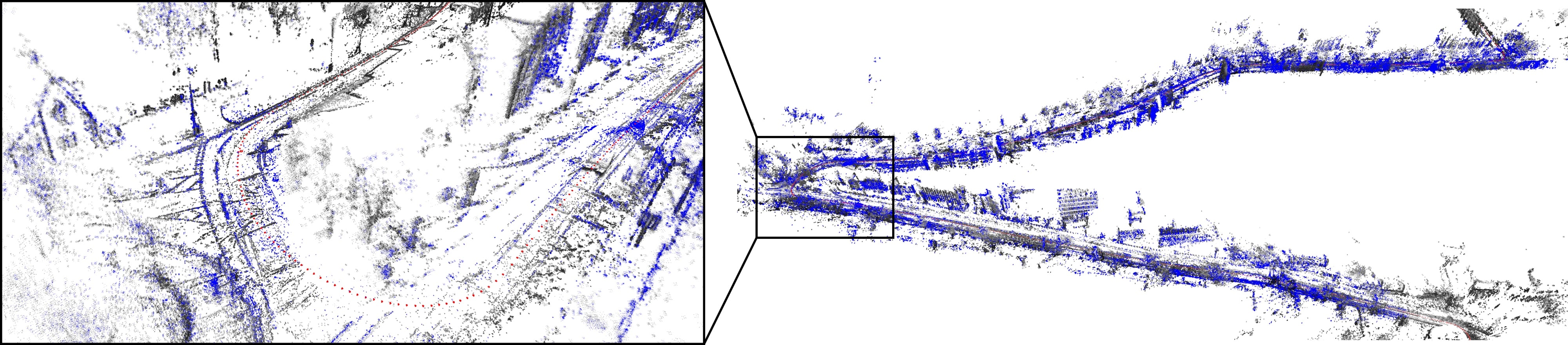}
    %\vspace{\capvspace}
    \caption{This figure shows a point cloud result of our GN-Net. We relocalize a rainy sequence (blue) against a reference map created from the sunny sequence (gray).}
    \label{fig:oxfordreloc}
    \vspace{\figvspace}
\end{figure*}

\begin{figure}[ht]
    \centering
    \includegraphics[width=0.9\linewidth]{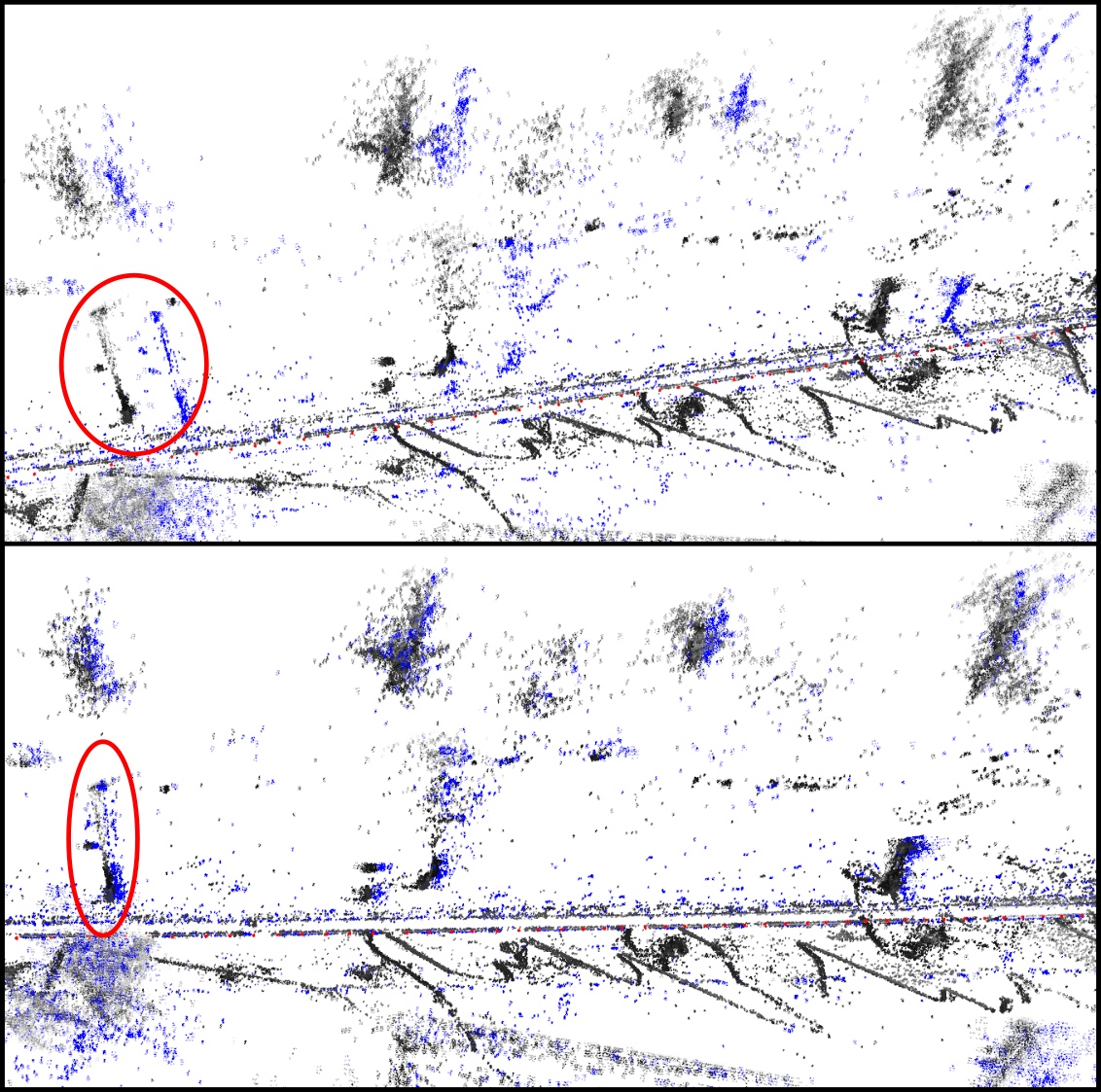}
    %\vspace{\capvspace}
    \caption{Top: relocalization using the model trained with only the contrastive loss. Bottom: relocalization using the model trained with our loss formulation. This visually demonstrates the influence of the Gauss-Newton loss.}
    \label{fig:oxfordrelocvisual}
    \vspace{\figvspace}
\end{figure}

Finally, we show a relocalization demo comparing our GN-Net to DSO. For this, we load a point cloud from a sequence recorded in the sunny condition and relocalize against sequences from rainy and snowy conditions. For each keyframe, we try to track it against the nearest keyframe in the map according to the currently estimated transformation between the trajectory and the map. Figure~\ref{fig:oxfordreloc} shows that the point clouds from the different sequences align nicely, despite belonging to different weather conditions. This experiment shows that our method can perform the desired operations successfully on a real-world application, including relocalization against unseen weather conditions. Figure~\ref{fig:oxfordrelocvisual} demonstrates the difference between our Gauss-Newton loss and the contrastive loss. This shows that the quantitative improvement has a visible effect on the application of relocalization. 
Figure~\ref{fig:sampleimages} shows sample images used in the qualitative relocalizations.

\subsection{Additional experiments on EuRoC and CARLA}
In the supplementary, we provide more evaluations on datasets with and without brightness variations. This includes relocalization tracking on the CARLA benchmark and visual odometry on the EuRoC~\cite{Burri2016Euroc} dataset. We show that also in these situations our deep features significantly outperform DSO and ORB-SLAM because of their robustness to large-baselines. On the EuRoC dataset, we improve the DSO performance by almost a factor of 2 for low-framerates.

%% file: sections/conclusion.tex
\section{Conclusion \& Future Work}

With the advent of deep learning, we can devise feature space encodings that are designed to be optimally suited for the subsequent visual SLAM algorithms.  More specifically, we propose to exploit the probabilistic interpretation of the commonly used Gauss-Newton algorithm to devise a novel loss function for feature space encoding that we call the Gauss-Newton loss. It is designed to promote robustness to strong lighting and weather changes while enforcing a maximal basin of convergence for the respective SLAM algorithm. Quantitative experiments on synthetic and real-world data demonstrates that the Gauss-Newton loss allows us to significantly expand the realm of applicability of direct visual SLAM methods, enabling relocalization and map merging across drastic variations in weather and illumination.